\documentclass{article}

\PassOptionsToPackage{dvipsnames}{xcolor}
\usepackage{iclr2026_conference,times}
\usepackage{amsmath,amsfonts,bm}
\usepackage{xcolor}

\def\eqref#1{equation~\ref{#1}}

\def\1{\bm{1}}

\DeclareMathAlphabet{\mathsfit}{\encodingdefault}{\sfdefault}{m}{sl}
\SetMathAlphabet{\mathsfit}{bold}{\encodingdefault}{\sfdefault}{bx}{n}

\newcommand{\E}{\mathbb{E}}

\newcommand{\R}{\mathbb{R}}

\makeatletter
\def\@maketitle{
  \vbox{\hsize\textwidth
    \centering
    {\LARGE\sc \@title\par}
    \ificlrfinal
      \lhead{}
      \def\And{\end{tabular}\hfil\linebreak[0]\hfil
               \begin{tabular}[t]{c}\bf\rule{\z@}{24pt}\ignorespaces}
      \def\AND{\end{tabular}\hfil\linebreak[4]\hfil
               \begin{tabular}[t]{c}\bf\rule{\z@}{24pt}\ignorespaces}
      \begin{tabular}[t]{c}\bf\rule{\z@}{24pt}\@author\end{tabular}
    \else
      \lhead{}
      \def\And{\end{tabular}\hfil\linebreak[0]\hfil
               \begin{tabular}[t]{c}\bf\rule{\z@}{24pt}\ignorespaces}
      \def\AND{\end{tabular}\hfil\linebreak[4]\hfil
               \begin{tabular}[t]{c}\bf\rule{\z@}{24pt}\ignorespaces}
      \begin{tabular}[t]{c}\bf\rule{\z@}{24pt}Anonymous authors\\Paper under double-blind review\end{tabular}
    \fi
    \vskip 0.3in minus 0.1in
  }
}
\makeatother

\usepackage{booktabs}
\usepackage{multirow}
\usepackage{graphicx}
\usepackage{subcaption}
\usepackage{xcolor}
\usepackage{microtype}
\usepackage{xurl}
\usepackage{hyperref}

\usepackage{enumitem}
\hypersetup{
    colorlinks=true,
    linkcolor=MidnightBlue,
    citecolor=MidnightBlue,
    urlcolor=MidnightBlue
}
\usepackage[capitalise,noabbrev,nameinlink]{cleveref}

\newcommand{\modelname}{Domyn-Small}

\title{\modelname: A European 10B Reasoning Language Model}

\author{
{\small\bfseries Simone Angarano, Francesco Bertolotti, Federico D'Ambrosio, Michele Resta, Alessandro Rognoni,} \\
{\small\bfseries Nicolò Ruggeri, Dario Salvati, Andrea Valenti, Alberto Veneri, Martin Cimmino} \\[4pt]
\normalfont Domyn \\
\normalfont \texttt{models@domyn.com}
}

\fancypagestyle{firstpage}{
  \fancyhf{}
  \renewcommand{\headrulewidth}{0.4pt}
  \fancyhead[L]{\includegraphics[height=0.7cm]{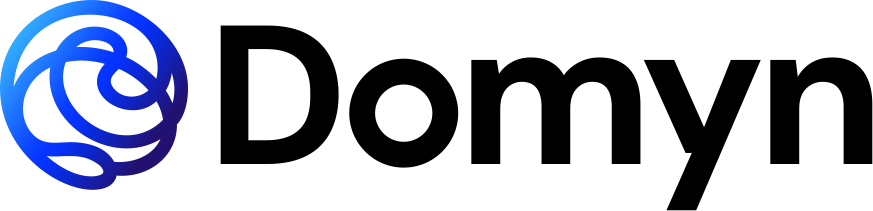}}
}

\iclrfinalcopy

\begin{document}

\thispagestyle{firstpage}
\maketitle

\begin{abstract}
We introduce \modelname{},\footnote{Weights available at \raisebox{-1pt}{\includegraphics[height=1em]{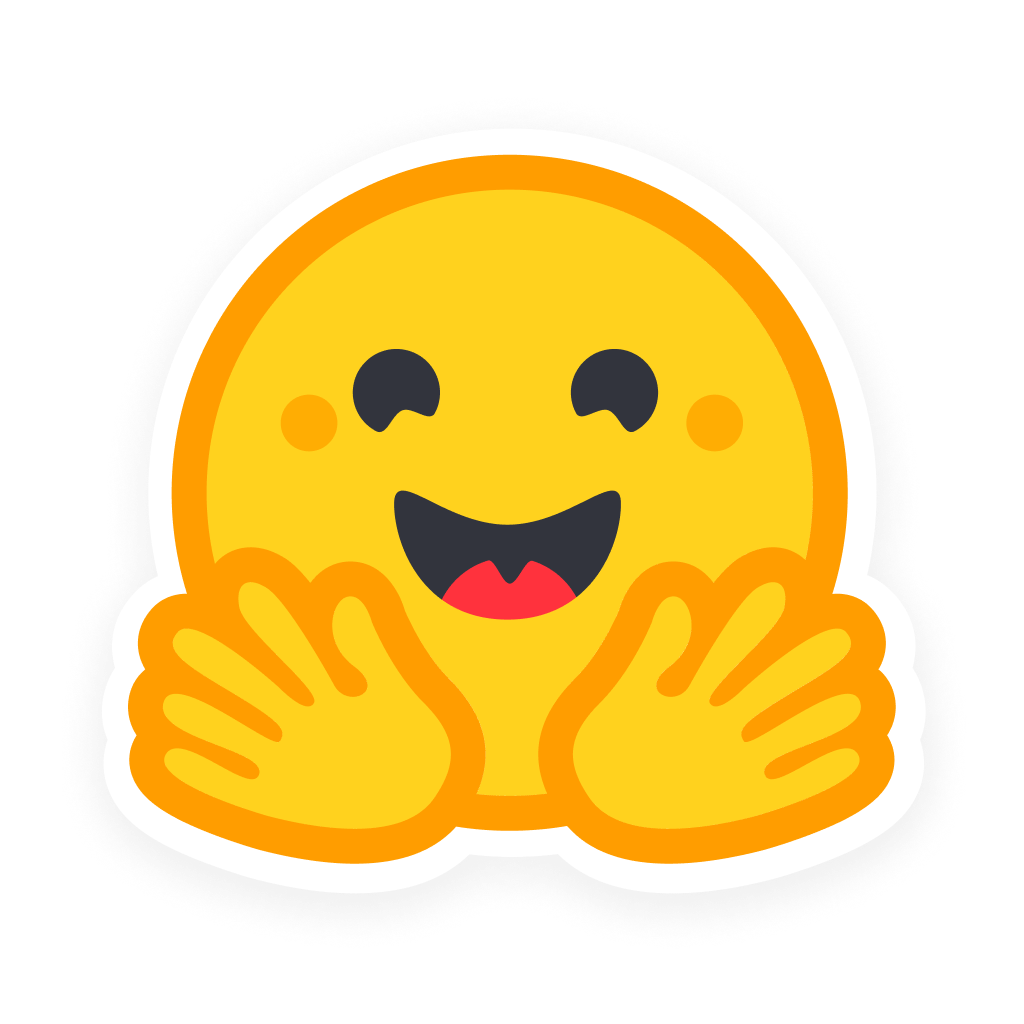}}\,\url{https://huggingface.co/domyn/Domyn-Small-v1.0}.} a 10-billion-parameter open-weight reasoning language model released under the MIT license.
\modelname{} is the product of an initial pre-training phase on 9 trillion tokens multilingual data, followed by a post-
training pipeline for reasoning, instruction following, and context extension. For the latter, we performed a Continued Pre-Training
(CPT) phase that doubles the native context window to 32K tokens, followed by SFT with a math-focused annealing run. Finally, the RL phase includes GRPO with verifiable rewards, DPO, and a multi-environment GRPO stage spanning five task domains: mathematics, code, multiple-choice QA, instruction-following, and tool calling.
The 32K-token native context extends to 128K at inference via YaRN, and a chat-template toggle enables dual-mode reasoning.
Against peer models in the 7--10\,B class (Qwen3.5-9B, OLMo-3-7B-Think, Nemotron-Nano-8B, Ministral-3-8B), \modelname{} achieves a strong accuracy-efficiency balance: it produces roughly one-third as many tokens as Qwen3.5-9B and approximately 35\,\% of OLMo-3-7B-Think's token budget on core reasoning benchmarks, while delivering strong instruction-following (IFEval 79.9) and competitive science reasoning (GPQA-Diamond 50.0).
We release the weights and the post-training recipe alongside Domyn Swarm (Apache~2.0), an open-source framework for scalable LLM inference on HPC clusters developed during this program and used throughout this work.

\end{abstract}

\section{Introduction}
\label{sec:introduction}

Regulated enterprises in financial services, defence, advanced manufacturing, and the public sector face a recurring constraint: they need capable language models that can be deployed and governed under European regulatory frameworks, including the EU AI Act~\citep{eu2024aiact}, sectoral data-residency rules, and procurement requirements that increasingly favour technology with auditable provenance.
However, these deployments operate under tight constraints.
Inference must run on infrastructure the enterprise controls, serving costs must be commensurate with high-volume agentic workloads, and the model must be compact enough to serve on a modest GPU allocation.
The 7--10\,B parameter class addresses this envelope directly: large enough to support multi-step reasoning, multilingual interaction, and tool use; small enough to serve at advantageous economics on a single GPU.

\begin{figure}[t]
  \centering
  \includegraphics[width=\textwidth]{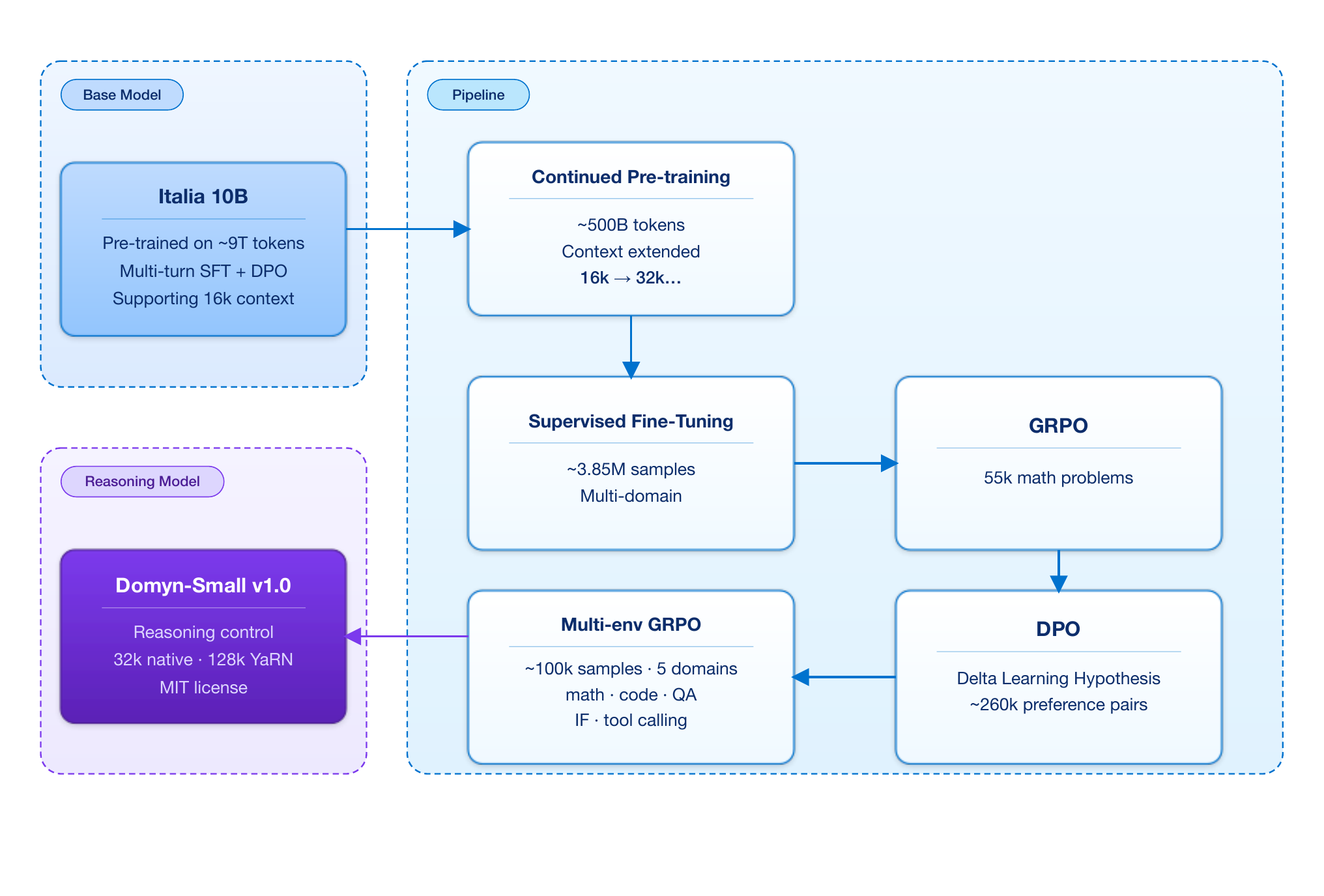}
  \caption{High-level training pipeline of \modelname{}. Starting from the Italia 10~B checkpoint, \modelname{} is the result of 5 adaptation stages encompassing CPT, SFT and multi-phase RL.}
  \label{fig:pipeline}
\end{figure}

\textbf{The \modelname{} model.}
To address this gap we introduce \modelname, a 10-billion-parameter open-weight reasoning language model released under the MIT license.
Rather than performing expensive pre-training, we chose to extend Italia~10B, a from-scratch foundation model trained in 2024 on approximately 9 trillion tokens and spanning 50+ languages. Italia~10B was trained as a non-reasoning model, unlike DeepSeek-R1~\citep{deepseekai2025r1} and the wave of open reasoning-focused recipes that followed. To improve the reasoning capabilities, we extend Italia~10B's strong foundation through a five-stage adaptation pipeline based on modern, state-of-the-art practices in context extension and post-training. \\
Our adaptation pipeline (\Cref{fig:pipeline}) runs end-to-end on the CINECA Leonardo Supercomputer~\citep{turisini2024leonardo} in partnership with CINECA and consists of:
\begin{itemize}
\item A Continued Pre-Training (CPT) phase that doubles the native context window from 16{,}384 to 32{,}768 tokens (\Cref{sec:pretraining:cpt})
\item A Supervised Fine-Tuning (SFT) on a 12.3-billion-token instruction mixture, including a short math-focused annealing run before reinforcement learning (\Cref{sec:posttraining:sft})
\item A Reinforcement Learning (RL) multi-stage phase, which we build with general capability in mind: \emph{i)} Group Relative Policy Optimisation (GRPO) under verifiable math rewards (\Cref{sec:posttraining:grpo}); \emph{ii)} Direct Preference Optimisation (DPO) under the Delta Learning Hypothesis (\Cref{sec:posttraining:dpo}) \emph{iii)} A final multi-environment GRPO stage spanning five task domains:mathematics, code, multiple-choice QA, instruction-following, and tool calling (\Cref{sec:posttraining:multienv}).
\end{itemize}
At inference, the 32{,}768-token native context extends to 131{,}072 tokens via YaRN~\citep{peng2023yarn}, i.e., four times the native context, and a chat-template toggle exposes a dual-mode reasoning capability (\emph{thinking on / thinking off}).

\textbf{Results.}
In \Cref{sec:evaluation} we compare against four peer models: Qwen3.5-9B~\citep{qwen2026qwen35}, OLMo-3-7B-Think~\citep{allenai2025olmo3}, and Llama-3.1-Nemotron-Nano-8B-v1~\citep{nvidia2025nemotronnano} from the 7--10\,B class, plus Ministral-3-8B-Reasoning-2512~\citep{mistral2025ministral3b} as an efficiency reference point.
\modelname{} achieves the best accuracy-efficiency balance in the 7--10\,B class, producing roughly one-third the tokens of Qwen3.5-9B on reasoning benchmarks (\Cref{sec:evaluation:efficiency}).
On accuracy, \modelname{} leads Nemotron-Nano and Ministral-3-8B-Reasoning-2512 on instruction-following and GPQA-Diamond, is at rough parity with Ministral-3-8B-Reasoning-2512 on AIME~2025, and trails Qwen3.5-9B and OLMo-3-7B-Think materially on hard-math reasoning (\Cref{sec:evaluation:capabilities}).
This reasoning gap reflects both the convergence state of the RL phase at checkpoint time and the 2024-vintage base model, which predates the wave of reasoning-optimised pre-training practices that underpin the strongest current competitors.

\textbf{Contributions.}
We make the following contributions:
\begin{itemize}
  \item \textbf{Model weights.} \modelname{} is released under the MIT license on Hugging Face, with the bf16 reference weights and a tokenizer compatible with the broader Domyn model family.
  \item \textbf{Open end-to-end post-training pipeline.} In the spirit of open science sharing (see e.g. OLMo~3~\citep{allenai2025olmo3}) we document the full adaptation pipeline from the Italia~10B foundation model through multi-environment GRPO at recipe-level fidelity (CPT data and RoPE base trajectory, SFT mixture composition, GRPO reward design and prompt set, DPO dataset, and multi-env GRPO hyperparameters and environment configuration).
  \item \textbf{Token-efficiency Pareto characterisation.} We present a systematic accuracy / mean-tokens-per-problem comparison against four peer models on three core reasoning benchmarks, establishing \modelname{}'s position on the deployment frontier and contextualising its efficiency advantage against both larger and smaller peers.
  \item \textbf{Domyn Swarm.} Domyn Swarm,\footnote{\url{https://github.com/igeniusai/domyn-swarm}; initial public release 6 October 2025.} an Apache~2.0 framework for scalable LLM-as-a-Service inference on HPC clusters, was developed during this program and open-sourced to support synthetic data generation and at-scale evaluation against teacher models on Leonardo (\Cref{sec:domynswarm}).
\end{itemize}

In summary, we release \modelname{} and its complete recipe to serve as a critical enabler for users and regulated industries operating at the intersection of effective model usage and efficient deployment needs.

\section{Model Architecture}
\label{sec:architecture}

\modelname{} is a dense, decoder-only causal Transformer~\citep{vaswani2017attention} with approximately 10 billion parameters, following established open decoder-only practice~\citep{touvron2023llama2,chowdhery2022palm}.
The model has 40 layers with hidden size 4{,}096 and uses Grouped-Query Attention~\citep{ainslie2023gqa} with 48 query heads and 8 key-value heads of 128-channel width to limit KV-cache bandwidth at the 32{,}768-token native context.
Position information is encoded through Rotary Position Embeddings~\citep{su2021roformer} applied to half of each head's feature dimensions, with the RoPE base set to 500{,}000 at the second-stage continued pre-training boundary (\Cref{sec:pretraining:cpt}); at inference, the context window is extended 4$\times$ to 131{,}072 tokens via training-free YaRN scaling~\citep{peng2023yarn}.
The MLP sub-layers use squared ReLU activations~\citep{so2021primer} with a 4$\times$ expansion to a 16{,}384-dimensional intermediate size and no gating.
Sub-layers are pre-normalised with LayerNorm~\citep{ba2016layernorm}; following standard practice in recent open models, biases are disabled in all linear and attention projections, input and output embeddings are untied, dropout is set to zero throughout training, and all weights are stored and trained in bfloat16.
The full hyperparameter specification is given in \Cref{tab:architecture}.

\begin{table}[!htbp]
  \caption{\modelname{} architectural specification.}
  \label{tab:architecture}
  \centering
  \begin{tabular}{lc}
    \toprule
    \textbf{Hyperparameter} & \textbf{Value} \\
    \midrule
    Layers                                    & 40 \\
    Hidden size ($d_{\mathrm{model}}$)        & 4{,}096 \\
    FFN intermediate size                     & 16{,}384 \\
    Attention heads                           & 48 \\
    KV heads (GQA)                            & 8 \\
    Head dimension                            & 128 \\
    Vocabulary size                           & 256{,}000 \\
    Native context length                     & 32{,}768 \\
    Extended context (YaRN, $4{\times}$)      & 131{,}072 \\
    RoPE base (post-CPT)                      & 500{,}000 \\
    MLP activation                            & squared ReLU \\
    Precision                                 & bfloat16 \\
    Total parameters                          & $\approx 10$\,B \\
    \bottomrule
  \end{tabular}
\end{table}

\textbf{Tokenizer.}
\modelname{} inherits Italia~10B's Byte-Pair Encoding tokenizer trained with SentencePiece~\citep{kudo2018sentencepiece}, with a 256{,}000-token vocabulary shared with Italia~10B's contemporary, the Colosseum~355B foundation model~\citep{nvidia2024colosseum}, for representational consistency across the model family.
The tokenizer covers 50+ languages with particular emphasis on European languages; following PaLM~\citep{chowdhery2022palm}, digits are split into individual digit tokens.

\section{Pre-Training}
\label{sec:pretraining}

The pre-training of \modelname{} proceeded in two stages.
The model descends from Italia~10B (\Cref{sec:pretraining:foundation}), a 10-billion-parameter base trained from scratch on approximately 9 trillion tokens at the end of 2024.
Italia~10B's 16{,}384-token native context was too short for long-form retrieval-augmented workloads, and its general-domain pre-training data mix lacked the concentration of technical, long-form content required for the reasoning workloads motivating this release.
We therefore inserted a dedicated second-stage long-form pre-training phase, which we refer to as continued pre-training (CPT). During CPT the model is exposed to 503 billion additional tokens of higher-quality, more technical content and the native context window is extended to 32{,}768 tokens (\Cref{sec:pretraining:cpt}).

\subsection{Foundation Pre-Training}
\label{sec:pretraining:foundation}

Italia~10B shares the architecture described in \Cref{sec:architecture}; the only architectural difference is that the RoPE base remained at 10{,}000 throughout foundation training and was raised to 500{,}000 only at the CPT boundary (\Cref{sec:pretraining:cpt}).

\textbf{Data composition.}
The pre-training corpus consists of curated documents drawn from web crawl, news, scientific papers, books, encyclopedic sources, source code, and a deliberately up-sampled multilingual long tail with emphasis on European languages.
The high-level token blend is approximately 60\,\% English natural language, 15\,\% multilingual natural language across 50+ languages with emphasis on European languages (German, French, Italian, Spanish, Portuguese, Russian, Romanian, Polish), 5\,\% mathematical content, and 20\,\% source code spanning 40+ programming languages.\footnote{We report macro-category percentages rather than source-level provenance.}

\textbf{Curation pipeline.}
The curation pipeline applied is made of four stages.
Document-level near-deduplication used MinHash-based approximate matching to reduce memorisation and increase diversity~\citep{lee2022deduplicating}.
Language-model quality filtering followed the CCNet methodology~\citep{wenzek2019ccnet}, retaining documents whose perplexity under a small reference language model fell within a calibrated band.
Heuristic filters extending the Gopher~\citep{rae2022gopher} and C4~\citep{raffel2020t5} recipes (symbol-to-word ratio, unique-word count, repetition fraction, document length) removed degenerate documents.
A final pass enforced personally identifiable information removal, copyright filtering against dynamic blocklists, and compliance with robots.txt and machine-readable opt-out signals as required by the EU Digital Single Market Directive Article~4(3)~\citep{eu2019dsm}.

\textbf{Compute and training schedule.}
Foundation training was executed in two phases on two clusters using the NeMo / Megatron-LM~\citep{shoeybi2019megatron} framework with tensor and context parallelism, and a distributed Adam optimiser~\citep{korthikanti2022activation}; both phases ran in bf16-mixed precision.
The first phase processed approximately 6 trillion tokens on NVIDIA DGX Cloud at a 4{,}096-token sequence length, where Italia~10B was developed alongside Domyn's larger Colosseum~355B foundation model~\citep{nvidia2024colosseum} on shared infrastructure and data pipelines.
The second phase processed the remaining approximately 3 trillion tokens on the CINECA Leonardo supercomputer using 512 NVIDIA A100-SXM-64\,GB GPUs; this phase implemented a progressive sequence-length curriculum, extending from 4{,}096 to 16{,}384 tokens.
At sustained per-GPU throughput comparable to the subsequent continued pre-training phase (\Cref{sec:pretraining:cpt}), and slightly higher owing to the shorter sequence length, the Leonardo foundation slice consumed approximately 417{,}000 GPU-hours at $\sim$38\,\% Model FLOPs Utilisation~\citep{chowdhery2022palm}.

\subsection{Continued Pre-Training: Second-Stage Long-Form Pre-Training}
\label{sec:pretraining:cpt}

We frame this stage as a second pre-training pass rather than as pure context extension based on two grounds.
First, its 503-billion-token budget exceeds typical context-extension budgets by an order of magnitude, and is more appropriately read as a deliberate continuation of pre-training~\citep{hoffmann2022chinchilla}.
Second, the data distribution is materially re-weighted relative to foundation training: the share of tokens allocated to mathematical reasoning and source code is substantially increased, and a tighter quality bar is applied across all categories, exposing the model not only to longer contexts but also a materially different and higher-quality distribution.

\begin{figure}[t]
  \centering
  \begin{subfigure}[t]{0.48\textwidth}
    \centering
    \includegraphics[width=\textwidth]{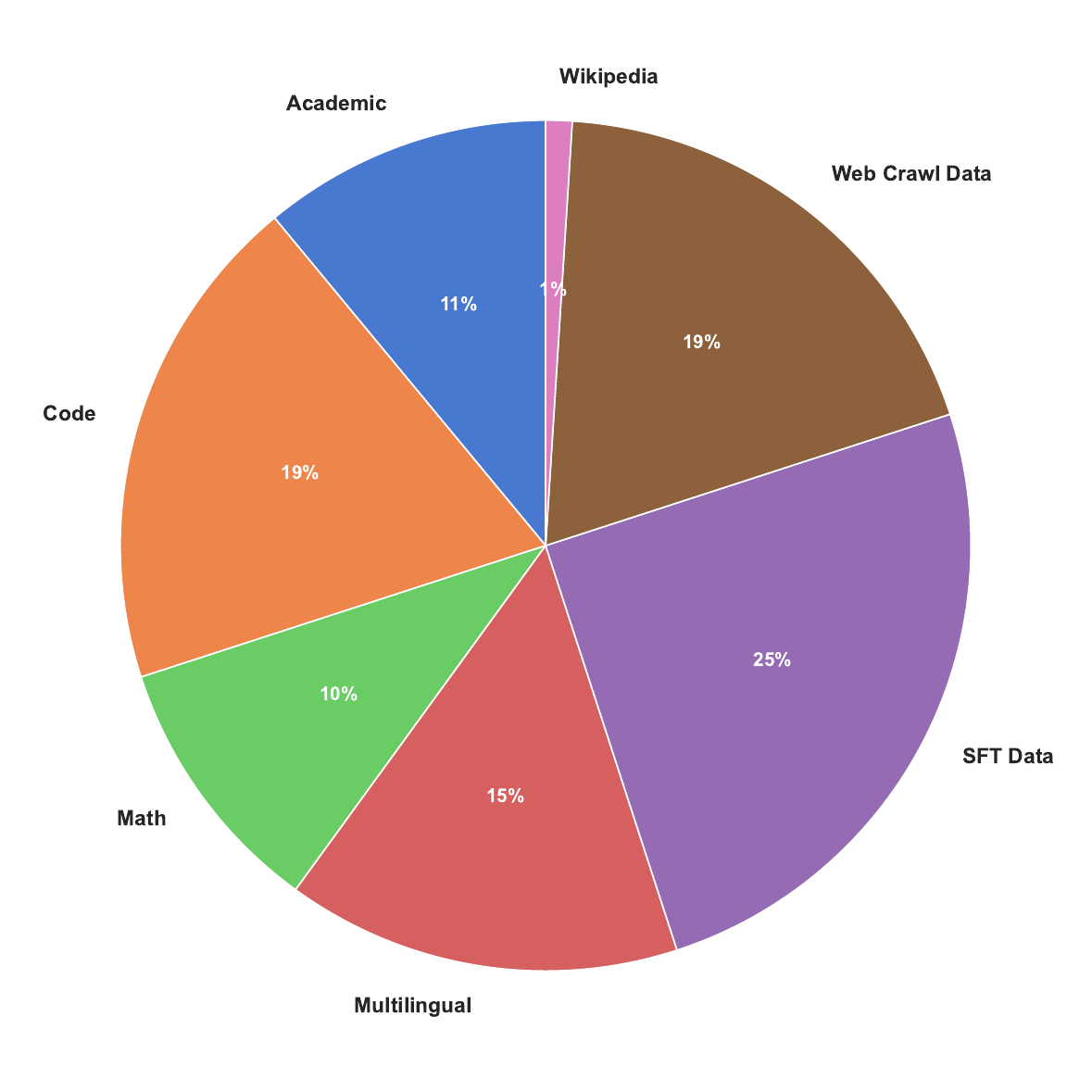}
    \caption{Data mix by category.}
    \label{fig:cpt:mix:sub}
  \end{subfigure}
  \hfill
  \begin{subfigure}[t]{0.48\textwidth}
    \centering
    \includegraphics[width=\textwidth]{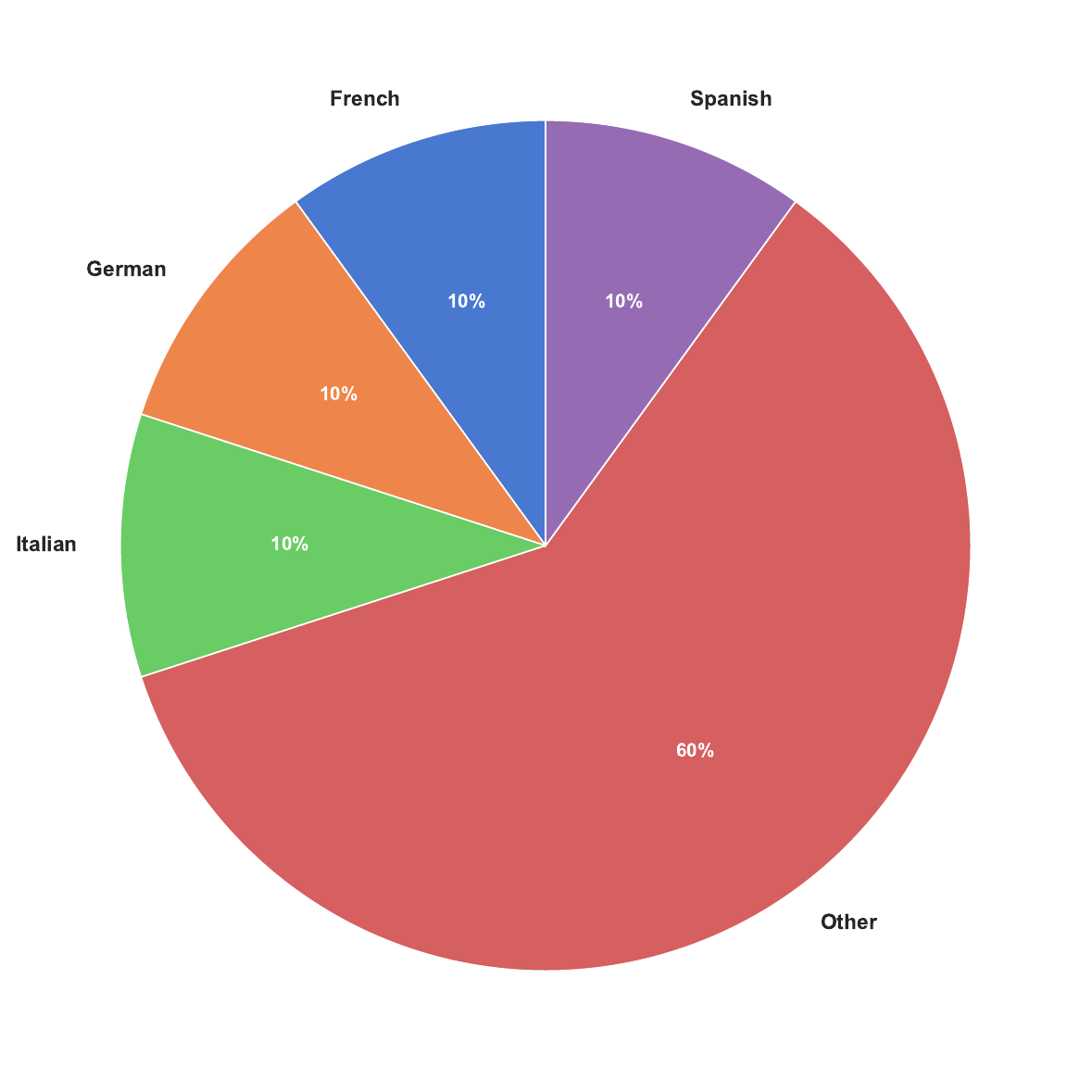}
    \caption{Per-language weighting within the multilingual stream.}
    \label{fig:cpt:multilingual:sub}
  \end{subfigure}
  \caption{Data composition across the 503B-token CPT corpus. \textbf{(a)}~Data mix by category. SFT-style instruction data is the largest single stream (25\,\%), followed by web crawl and code at $\sim$19\,\% each. The CPT mix is shared with the pre-training corpus of Domyn~Large~\citep{bertolotti2026domynlarge}. \textbf{(b)}~Per-language weighting within the multilingual stream. Tier~A languages (Spanish, Italian, French, German) each receive 10\,\% of the multilingual sampling budget; the remaining 60\,\% is allocated across the lower tiers, covering the bulk of the 56 languages drawn from FineWeb-2.}
  \label{fig:cpt:data}
\end{figure}

\textbf{Data composition.}
The aim of the CPT phase is to adapt the model to a re-weighted linguistic distribution by exposing it to large volumes of higher-quality, longer-form, and more technical content than the Italia~10B foundation mix, as well as to enrich the model's internal representations to capture language-specific morphology, syntax, semantics, and idiomatic patterns that are underrepresented in English-centric web crawls.
The CPT data mix is shared with the pre-training corpus of Domyn~Large~\citep{bertolotti2026domynlarge}, and is organised at the category level rather than as a flat union of sources: data was cleaned, filtered, deduplicated, partitioned into broad streams, and sampled from according to a predefined blend rather than to raw token availability.
Roughly 3~trillion tokens of source material were organized into the streams summarised in \Cref{fig:cpt:mix:sub}, of which the 503~billion top-quality ones were eventually retained for training.
The streams comprise: web-crawl content from DCLM~\citep{li2024dclm} and Dolma~\citep{soldaini2024dolma}, which together supply the bulk of the natural-language tokens at quality levels selected for second-stage training; source code from The Stack~v2~\citep{lozhkov2024stackv2}, retaining the multi-language code coverage of the foundation mix at higher token share (the subset used was filtered to retain only files carrying an explicit permissive license header); SFT-style instruction data, the largest single slice of the mix at 25\,\%; mathematical content from Nemotron-CC-Math~\citep{su2024nemotroncc}, increasing the share of step-by-step mathematical derivations relative to foundation training and exposing the model to longer chains of reasoning at the target sequence length; long-form scientific writing from ArXiv and peS2o~\citep{soldaini2023pes2o}, which taken together supply the academic documents that motivate the 32{,}768-token native context; encyclopedic coverage from Wikipedia; and a multilingual stream drawn from the high-quality subset of FineWeb-2~\citep{penedo2024fineweb} across 56 languages.

\textbf{Multilingual stream and tier weighting.}
Within the multilingual stream we explicitly control the sampling probability across the 56 FineWeb-2 high-quality languages: languages are partitioned into four tiers (A, B, C, and an ``Others'' bucket), and sampling probability decays from Tier~A through to ``Others''.
Tier~A places Spanish, Italian, French, and German at the top of the schedule with elevated weight at 10\,\% each of the multilingual stream, and the remaining 60\,\% is distributed across the lower tiers (\Cref{fig:cpt:multilingual:sub}).
This is a deliberate prioritisation of European languages, consistent with the broader sovereign-EU framing of the project.

\textbf{Context-window extension and RoPE rescaling.}
At the CPT boundary, the RoPE base frequency is increased from 10{,}000 to 500{,}000~\citep{su2021roformer}, an adjustment that rescales the rotary embeddings to encode the longer relative positions the extended window introduces, without re-initialising the position-encoding modules~\citep{xiong2023effective}.
The phase trains end-to-end at the target sequence length of 32{,}768 tokens. At inference time, the 32{,}768-token native context is further extended to 131{,}072 tokens via YaRN~\citep{peng2023yarn}, a training-free 4$\times$ RoPE-scaling method whose 500{,}000 RoPE base is the supported regime.

\textbf{Compute and training schedule.}
The CPT phase ran on the CINECA Leonardo supercomputer using 256 NVIDIA A100-SXM-64\,GB GPUs under the NeMo / Megatron-LM~\citep{shoeybi2019megatron} framework, with tensor parallelism 4 inside a node, context parallelism 2, and sequence parallelism enabled; the phase ran in bf16-mixed precision throughout, with a distributed Adam optimiser~\citep{korthikanti2022activation}.
The data-parallel replica count was 32, with micro-batch size 1 and a global batch size of 128 sequences, giving 4.19 million tokens per optimiser step at the 32{,}768-token sequence length.
Optimisation used Adam ($\beta_1=0.9$, $\beta_2=0.95$, $\epsilon=10^{-5}$, weight decay 0.1, gradient clip 1.0) on a cosine schedule that decayed from a peak learning rate of $2 \times 10^{-5}$ to $1 \times 10^{-6}$ over 120{,}000 steps, after a 500-step linear warmup.
The peak learning rate is approximately fifteen times below a typical fresh-pre-training peak for a 10B-parameter model, the standard ratio for limiting catastrophic forgetting during continued pre-training.
The run consumed approximately 77{,}000 GPU-hours at $\sim$35\,\% Model FLOPs Utilisation~\citep{chowdhery2022palm} and processed 503 billion training tokens.

\section{Post-Training}
\label{sec:posttraining}

The post-training pipeline took the post-CPT base model produced in~\Cref{sec:pretraining:cpt} and converted it into the released \modelname{}-v1.0 instruction model through four sequential stages.
First, a long supervised fine-tuning pass over a multi-task instruction mixture, followed by a short math-focused annealing run that sharpened the mathematical-reasoning before reinforcement learning (\Cref{sec:posttraining:sft}).
Second, Group Relative Policy Optimisation under verifiable rewards on a math-only prompt set (\Cref{sec:posttraining:grpo}).
Third, a Direct Preference Optimisation pass under the Delta Learning Hypothesis (\Cref{sec:posttraining:dpo}).
Fourth, a multi-environment GRPO stage that extended reinforcement learning across five task domains simultaneously, producing the v1.0 release weights (\Cref{sec:posttraining:multienv}).
\subsection{Supervised Fine-Tuning}
\label{sec:posttraining:sft}

The supervised fine-tuning (SFT) phase aimed to enhance \modelname{}'s instruction-following and reasoning capabilities. To do so, we built a comprehensive multi-task mixture of publicly released instruction datasets, covering a wide range of tasks and languages.
The mixture was designed to support dual-mode (on/off) reasoning and was supported by a custom chat template that was preserved across all downstream stages.
The stage started from the CPT checkpoint of~\Cref{sec:pretraining:cpt} (with a context length of 32{,}768 tokens) and ran for approximately 1.13 epochs over a multi-task mixture of 3.85M samples (12.29B tokens). We maintained the same architecture, tokenizer, and context length throughout the whole post-training pipeline, so that all subsequent stages were initialised from and directly comparable to the SFT checkpoint described here.

\textbf{Mixture composition.}
The training mixture was assembled from more than 40 publicly released instruction datasets through a four-stage pipeline: 
\begin{itemize}
\item \textit{format verification} to ensure sample correctness and consistency;
\item \textit{language filtering} on prompts, answers, and reasoning traces based on pre-defined language tiers;
\item \textit{length filtering} capped at 32{,}600 tokens to leave headroom for chat-template tokens;
\item \textit{conversion} to the NeMo chat format.
\end{itemize}
After filtering and carefully weighting source datasets, the mixture contained 3{,}845{,}862 samples and 12.29\,B tokens, 10.40\,B of which were loss-bearing (assistant-side) tokens. The samples were distributed across 10 task categories and 57 languages.
Sub-sampling was applied most aggressively to the largest reasoning-style data sources, while smaller, editorially curated sources were retained at full weight.
The resulting composition of the main task categories is given in~\Cref{tab:sft-mixture}.

\begin{table}[!htbp]
  \caption{SFT mixture composition by task category. \textit{\% gen tokens} is the share of loss-bearing (assistant-side) tokens.}
  \label{tab:sft-mixture}
  \centering
  \small
  \begin{tabular}{lrrrr}
    \toprule
    \textbf{Task} & \textbf{Samples} & \textbf{\% samples} & \textbf{\% tokens} & \textbf{\% gen tokens} \\
    \midrule
    STEM                  & 824{,}478 & 21.4 & 16.9 & 17.9 \\
    Function calling      & 543{,}748 & 14.1 & 5.9  & 1.9  \\
    Math                  & 538{,}533 & 14.0 & 20.4 & 23.7 \\
    Code                  & 522{,}158 & 13.6 & 21.1 & 22.8 \\
    Instruction following & 462{,}915 & 12.0 & 3.8  & 2.3  \\
    Multilingual          & 430{,}884 & 11.2 & 18.7 & 21.3 \\
    Chat                  & 389{,}616 & 10.1 & 8.5  & 9.1  \\
    Long context          & 78{,}215  & 2.0  & 4.3  & 0.5  \\
    Safety                & 51{,}839  & 1.4  & 0.3  & 0.3  \\
    \bottomrule
  \end{tabular}
\end{table}

\textbf{Dual reasoning.}
The mixture was partitioned into \textit{reasoning-on} samples carrying explicit \texttt{<think>$\ldots$</think>} traces, and \textit{reasoning-off} samples that respond directly without an intermediate trace (see~\Cref{tab:sft-mixture-reasoning} for a breakdown).
At inference time, the two modes are selected via system-level instructions \texttt{thinking on} and \texttt{thinking off}. The toggle was implemented purely at the chat-template level, following the dual-mode framing adopted by other recent reasoning models~\citep{deepseekai2025r1,yang2025qwen3}.
The per-task reasoning balance was tuned to match each category's intended downstream use. For instance, Math and Code splits contain a high percentage of reasoning samples due to their complexity.
On the other hand, Function calling (15\,\% reasoning-on) and Instruction following (23\,\% reasoning-on) were deliberately reasoning-off-dominant to preserve the strictness their downstream applications require.

\begin{table}[!htbp]
  \caption{SFT mixture composition by reasoning mode. \textit{\% gen tokens} is the share of loss-bearing (assistant-side) tokens.}
  \label{tab:sft-mixture-reasoning}
  \centering
  \small
  \begin{tabular}{lrrrr}
    \toprule
    \textbf{Reasoning mode} & \textbf{Samples} & \textbf{\% samples} & \textbf{\% tokens} & \textbf{\% gen tokens} \\
    \midrule
    ON  & 2{,}070{,}827 & 53.8 & 83.3 & 92.4 \\
    OFF & 1{,}775{,}035 & 46.2 & 16.7 & 7.6  \\
    \bottomrule
  \end{tabular}
\end{table}

\textbf{Language coverage.}
The mixture was dominated by English samples (78.5\,\% of tokens), but retained a substantial coverage of European languages: a Tier-A bucket of Italian, German, French, and Spanish accounted for 20.2\,\% of tokens, while Tier-B languages (e.g., Chinese, Portuguese, Russian) covered 1.1\,\% of them. A Tier-C long tail covered the remaining 0.2\,\%. The detailed distribution for the top 10 languages can be found in~\Cref{tab:top_languages_by_tokens}.

\begin{table}[htbp]
\caption{Top 10 languages by token count.}
\label{tab:top_languages_by_tokens}
\centering
\begin{tabular}{lrr}
\toprule
Language & Tokens (M) & Percentage (\%) \\
\midrule
English & 9,643 & 78.50 \\
Italian &   781 &  6.36 \\
German &   596 &  4.85 \\
French &   585 &  4.76 \\
Spanish &   516 &  4.20 \\
\midrule
Chinese &    97 &  0.79 \\
Portuguese &    19 &  0.15 \\
Russian &    14 &  0.11 \\
Tamil &     4 &  0.03 \\
Arabic &     3 &  0.03 \\
\bottomrule
\end{tabular}
\end{table}

\textbf{Training setup.}
SFT ran on 128 Leonardo nodes (512 NVIDIA A100-SXM-64\,GB GPUs) using the NeMo framework with a Tensor Parallelism (TP) of 4, a Context Parallelism (CP) of 4, and BF16 mixed precision. With this setup and a gradient accumulation factor of 4, we obtained a global batch size (GBS) of 128 sequences across the 32 data-parallel replicas. This led to 4.19M tokens per optimiser step.
Adam was used as the distributed optimiser~\citep{korthikanti2022activation} ($\beta_1=0.9$, $\beta_2=0.999$, $\epsilon=10^{-8}$) with a weight decay of 0.01 and gradient clipping at 1.0. 
The learning rate scheduler was a 1{,}000-step linear warmup followed by a flat learning rate of $1 \times 10^{-5}$.
This flat-LR regime is a deliberate departure from the cosine-decay schedule used during foundation and CPT training, intended to avoid late-training under-shooting on a long, multi-task SFT mixture where the loss continued to decrease throughout the run.

\paragraph{Math-focused annealing.}
Following the long SFT pass we ran a short, math-focused annealing stage to sharpen the model's mathematical-reasoning prior before reinforcement learning, modelled on similar pre-RL annealing stages used in recent reasoning-focused models~\citep{deepseekai2025r1}.
The annealing run was initialised from the released SFT checkpoint and trained on a math-heavy slice of the SFT mixture for approximately 3{,}000 optimiser steps. The same distributed configuration, optimiser settings, and 32{,}768-token sequence length were kept from the previous stage.
\subsection{Group Relative Policy Optimisation with Verifiable Rewards}
\label{sec:posttraining:grpo}

After SFT and the math-focused annealing stage, we further trained the model with Reinforcement Learning under Verifiable Rewards (RLVR), using Group Relative Policy Optimisation (GRPO)~\citep{shao2024deepseekmath}.
Following the DeepSeekMath / DeepSeek-R1 RLVR formulation~\citep{deepseekai2025r1}, the reward signal was supplied entirely by rule-based verifiers operating on the model's output.
In this stage, we restricted the GRPO training to mathematical reasoning tasks.

\textbf{Initialisation and prompt set.}
The GRPO run was initialised from the final SFT stage described in \Cref{sec:posttraining:sft}.
The training prompt set comprised 55{,}373 math problems taken from the DeepScaleR data blend~\citep{deepscaler2025}. Some samples were duplicated with different system prompts to create both a reasoning-on and a reasoning-off version of the same problem, for a final ratio of 80\%/20\% of reasoning-on/reasoning-off problems, respectively.
A held-out validation set was used for evaluation after every 50 optimizer steps for monitoring, while the final benchmark evaluation was performed on a standalone evaluation harness (\Cref{sec:evaluation}).

\textbf{Reward function.}
The chosen reward consisted of the additive combination of three components.
\begin{itemize}
\item Mathematical correctness, checked by a symbolic verifier returning $\{1, 0, -0.8\}$ for math-correct, math-incorrect, and verifier-error outcomes respectively, and was then scaled by $\times 3$ so that the math-correctness component spanned $[-2.4, +3]$.
\item A binary format component checked for the presence of exactly one \texttt{</think>} closer (in reasoning-on mode) or its absence (in reasoning-off mode), and for exactly one answer within \texttt{\textbackslash boxed\{$\cdot$\}} in the post-thinking segment. The format component was the average of the two checks and contributed at most $1.0$.
\item A soft length penalty applied to the number of generated tokens, decreasing linearly from $0$ to $-0.25$ at the 8{,}192-token maximum response length.
\end{itemize}
The aggregate reward range was approximately $[-3, +4]$.

\textbf{Algorithm and hyperparameters.}
We used group-normalised advantages with group size $G = 8$ rollouts per prompt, mean-centred and standardised by the within-group standard deviation.
We did not use the KL loss penalty term in our trainings, as we noted that its inclusion decreased performance without improving stability.
We used an advantage clipping range of $\epsilon = [0.2, 3.0]$. Training was done on-policy with one gradient pass per rollout batch.
Optimisation used Adam ($\beta_1 = 0.9$, $\beta_2 = 0.999$) with weight decay $0.01$, gradient clipping at $1.0$, and a constant learning rate of $1 \times 10^{-6}$.

\textbf{Distributed setup and schedule.}
We trained in a distributed setting of 64 A100 GPUs, with half the GPUs running a vLLM~\citep{kwon2023efficient} rollout engine in bfloat16 and half running an FSDP2-sharded actor update.
Each rollout step generated $128 \times 8 = 1{,}024$ trajectories. These trajectories were used to construct the GRPO mini-batches of size 128.
Decoding used temperature $1.0$, top-$p = 1$, a 4{,}096-token prompt cap, and an 8{,}192-token response cap (combined within the 16{,}384-token vLLM window).
The run was carried on for one epoch over the dataset.
At end of training, the model reached a mean reward score of 1.55 on the in-loop training distribution and 100\,\% format-success on the held-out math probe.

\subsection{Direct Preference Optimisation}
\label{sec:posttraining:dpo}

We conducted an additional post-training stage using Direct Preference Optimisation~\citep{rafailov2024dpo}, a preference-based fine-tuning algorithm that optimises the policy directly against a preference dataset without an intermediate reward model.
We chose to rely on this algorithm rather than RLHF-style PPO with a learned reward model for two reasons: first, to preserve the simplicity and verifiability of a pure rule-based reward signal, and second to put to test the Delta Learning Hypothesis~\citep{geng2025delta} on our model.

Geng et al.~argue that the quality of preference data depends primarily on the \emph{quality delta} between the chosen and the rejected response, not on the absolute quality of either side: pairing a high-capability response with a deliberately weak alternative therefore yields a useful contrastive signal even where supervised fine-tuning on the chosen response alone would not. We remand the interested reader to the original paper for the full argument and empirical support.
The dataset employed for this phase was the \textit{Dolci}\footnote{\url{https://huggingface.co/datasets/allenai/Dolci-Instruct-DPO}} dataset by AllenAI, released concurrently with OLMo 3~\citep{allenai2025olmo3}.

\textbf{Dataset.}
\textit{Dolci} comprises approximately 260{,}000 preference pairs assembled from three sources, each of which embeds a deliberate quality delta:
\begin{enumerate}[nosep,leftmargin=*]
  \item Heuristic pairs with chosen completions decoded from Qwen~3 32B and rejected completions from Qwen~3 0.6B~\citep{yang2025qwen3}.
  \item GPT-4.1-judged pairs in which GPT-4.1 selected the best response from a pool of candidates that always included a deliberately weak one, taking the worst response as rejected to maximise the delta.
  \item 10{,}000 multi-turn preference pairs: 5{,}000 with synthetic context, where related questions or paraphrases were prepended as prior turns to simulate a conversation history, and 5{,}000 with self-talk in which the LLM extended the initial prompt by generating its own follow-up requests.
\end{enumerate}

To mitigate length bias (the preference of LLMs to select significantly longer responses as correct and mark as rejected the shorter ones) the chat and multi-turn data was filtered to ensure that the chosen-vs-rejected length difference was less than 100 tokens.

\textbf{Training.}
The policy and the reference were both initialised from the post-GRPO checkpoint that follows \Cref{sec:posttraining:grpo}, with the reference held identical to the initialisation throughout training.
We trained with $\beta = 0.1$, peak learning rate $5 \times 10^{-7}$ with a cosine schedule with 10\,\% warmup, and the standard token-level DPO loss (i.e.\ not the length-normalised variant of~\citet{lambert2024tulu3}).
Training ran on 32 NVIDIA A100-SXM-64\,GB GPUs under FSDP in bf16-mixed precision, with per-device batch size 1 and gradient accumulation 64, giving an effective global batch size of 2{,}048 prompts per optimiser step.
The maximum prompt length is 640 tokens and the maximum total sequence length 3{,}456 tokens, sized from a length analysis of the full preference set against the \modelname{} tokenizer. The training run was configured to run for a maximum of two epochs over the 260{,}000-pair dataset, and it was early-stopped using a validation set benchmark.

\subsection{Multi-Environment GRPO}
\label{sec:posttraining:multienv}

The final post-training stage extends GRPO to five task domains simultaneously, producing the released \modelname{}-v1.0 weights.
Initialised from the DPO checkpoint of \Cref{sec:posttraining:dpo}, this stage is trained with NeMo RL v0.5.0~\citep{nvidianemorl2025}, a scalable asynchronous RL framework developed by NVIDIA, and orchestrated via NeMo Gym.
NeMo Gym\footnote{\url{https://github.com/NVIDIA-NeMo/Gym}} is the multi-environment layer that routes each prompt to the appropriate reward server, manages asynchronous tool-call execution loops for agentic tasks, and returns scalar rewards to the training loop.

\textbf{Environments and verifiers.}
Five NeMo Gym environments are active simultaneously:
\begin{itemize}
\item \textit{Mathematical reasoning} used the same rule-based symbolic verifier as \Cref{sec:posttraining:grpo}; the LLM judge was disabled so that only the rule-based signal was present.
\item \textit{Code generation} checked generated programs against unit tests executed in 1{,}024 sandboxed processes, with a per-test timeout of 10\,s, returning a binary pass/fail.
\item \textit{Multiple-choice question answering} was evaluated by exact-match against the ground-truth answer key.
\item \textit{Instruction-following} was evaluated by rule-based checkers.
\item \textit{Tool calling and agentic tasks} were evaluated by environment-grounded success signals over multi-turn trajectories of up to five turns.
\end{itemize}
All five environments return binary rewards in $\{0, 1\}$.

\textbf{Dataset.}
The training mixture was derived from the publicly released \texttt{nvidia/Nemotron-3-Nano-RL-Training-Blend}~\citep{nvidia2025nemotron3nanodata}.
The blend covered all five task domains; we applied domain re-weighting relative to the default Nemotron-3-Nano distribution to assign more mass to instruction-following and mathematics, the two domains with the most headroom in the DPO checkpoint.
Similarly to \Cref{sec:posttraining:grpo}, for every task type, 80\,\% of prompts were constructed in reasoning-on mode, while 20\,\% were left in reasoning-off mode, uniformly across all domains.

\textbf{Dynamic sampling and zero-gradient filtering.}
A fundamental challenge in GRPO rewards was group collapse: when all $G = 16$ rollouts in a group received the same reward (all correct or all incorrect), the within-group normalised advantage was identically zero and the group contributed no gradient signal.
In a multi-environment binary-reward setting this was especially acute at the difficulty extremes, where the model consistently failed or consistently succeeded on an entire group for either very-easy or very-hard prompts.
We addressed this with dynamic sampling: at each training step, the sampler generated up to ten generation batches and discarded groups whose reward variance was zero, assembling the effective training batch only from groups with informative contrastive signal.
This kept every gradient update free of degenerate zero-signal trajectories at the cost of variable generation overhead per step, bounded by ten generation batches.

\textbf{Training objective.}
Advantages were computed using GRPO with a leave-one-out baseline~\citep{shao2024deepseekmath,ahmadian2024rloo}: for each rollout $k$ in a group of $G = 16$, the baseline is the mean reward of the remaining $G{-}1$ rollouts, and advantages are normalised by the unbiased within-group standard deviation.
Groups where all rollouts receive the same reward have zero variance and are discarded by dynamic sampling before the gradient update.
The policy gradient loss follows the DAPO formulation~\citep{yu2025dapo} as implemented in NeMo RL~\citep{nvidianemorl2025}: a token-level clipped surrogate with asymmetric clip bounds $(\varepsilon_{\mathrm{low}}, \varepsilon_{\mathrm{high}}) = (0.2, 0.28)$, a dual-clip ceiling $c = 10$ that prevents over-suppression on strongly negative advantages, a token-level importance-sampling correction for the off-policy gap introduced by asynchronous generation, and a KL penalty $\beta = 0.01$ computed with the K3 estimator~\citep{schulman2020klblog}.

\textbf{Asynchronous rollout architecture.}
Training ran on 80 Leonardo nodes (320 NVIDIA A100-SXM-64\,GB GPUs): 16 nodes (64\,GPUs) were dedicated to asynchronous vLLM-based rollout generation, and the remaining 64 nodes (256\,GPUs) ran the DTensor/FSDP2-style policy update with context parallelism $\mathrm{CP}=2$ and full-parameter updates.
Generation and training overlapped asynchronously: the vLLM engine began sampling the next batch while the current batch's gradient update was in flight.
A near-on-policy constraint was enforced by capping trajectory age at one optimiser step; updated policy weights were pushed to the generation replicas immediately after each training step, invalidating and regenerating the vLLM KV cache.
The group size was $G = 16$ rollouts per prompt, with 32 prompts per step giving 512 trajectories per gradient update.

\textbf{KL regularisation and learning-rate stability.}
Unlike the math-only GRPO stage, which operated with KL disabled, this stage applied a KL penalty in the loss with coefficient $\beta = 0.01$ (K3 estimator).
A non-zero KL was necessary to prevent the policy from oscillating between task-specific optima; $\beta = 0.01$ achieved stable per-environment reward curves.
The learning rate was similarly empirically determined. The stable configuration was found at a constant learning rate of $1.5 \times 10^{-6}$, after a 30-step linear warmup, with AdamW ($\beta_1 = 0.9$, $\beta_2 = 0.999$, $\varepsilon = 10^{-8}$, weight decay $0.01$).

\textbf{PPO surrogate.}
We use DAPO-style asymmetric clipping: a lower clip at $\varepsilon_{\mathrm{low}} = 0.2$ and a wider upper clip at $\varepsilon_{\mathrm{high}} = 0.28$, which loosens the constraint on probability-ratio increases relative to symmetric clipping, thus encouraging exploration on positive-advantage trajectories.
A dual-clip ceiling of $c = 10$ prevents over-suppression on strongly negative advantages.
The combination of asymmetric clipping and dual-clipping produces a more exploration-friendly surrogate than the standard PPO loss, which is beneficial in the early phases of multi-environment RL where the policy must simultaneously improve across five distinct task distributions.

\textbf{Training schedule.}
The training is carried on for one epoch over the dataset.
Validation is run every 10 steps against a held-out split, and the top-3 checkpoints by validation accuracy are retained; the released v1.0 weights are the best-validation checkpoint from this run.

\subsection{Progressive Evaluation Across Post-Training Stages}
\label{sec:posttraining:progressive}

We complement the methodological description above with a within-pipeline view of how \modelname{}'s benchmark behaviour evolves between consecutive post-training stages, isolating which stage contributes which capability gain, including the regressions we observe on individual metrics.
The tables below trace the pipeline across all four post-training stages through to the released v1.0 weights; comprehensive per-benchmark evaluation appears in \Cref{sec:evaluation}.
All numbers below are taken from the same standalone evaluation harness used in \Cref{sec:evaluation}, run against the released checkpoints at the end of each intermediate stage.

\Cref{tab:posttraining-math-rl} isolates the effect of the math-only GRPO stage across ten benchmarks spanning mathematics, science reasoning, and multilingual arithmetic, where verifiable-reward RL has the most direct mechanism of action.
The largest gains concentrate on the in-domain MMLU College Mathematics ($+15.0$) and MMLU-Pro Math ($+12.8$) splits, with clear positive transfer onto MGSM, most strikingly on the German split ($+10.0$), and a more modest but consistent lift on GPQA-Diamond ($+3.5$).
The single regression is MATH-500 ($-2.6$), however it was recovered at the subsequent DPO stage (\Cref{tab:posttraining-overall}).

\begin{table}[!htbp]
  \caption{Math- and STEM-leaning benchmark scores at the SFT checkpoint and after GRPO with verifiable rewards on the math-only prompt set. All values are percentage accuracy. The right-most column reports the GRPO-induced delta.}
  \label{tab:posttraining-math-rl}
  \centering
  \small
  \begin{tabular}{lrrr}
    \toprule
    \textbf{Benchmark} & \textbf{SFT} & \textbf{+ GRPO (math-only)} & $\boldsymbol{\Delta}$ \\
    \midrule
    GPQA-Diamond                    & 35.86 & 39.39 & $+3.53$  \\
    MMLU College Mathematics        & 77.00 & 92.00 & $+15.00$ \\
    MMLU Abstract Algebra           & 80.00 & 89.00 & $+9.00$  \\
    MMLU Elementary Mathematics     & 94.44 & 96.56 & $+2.12$  \\
    MMLU-Pro Math                   & 68.76 & 81.57 & $+12.81$ \\
    MGSM-en                         & 92.40 & 94.80 & $+2.40$  \\
    MGSM-de                         & 78.40 & 88.40 & $+10.00$ \\
    MGSM-fr                         & 77.20 & 80.00 & $+2.80$  \\
    MGSM-es                         & 82.40 & 85.60 & $+3.20$  \\
    MATH-500                        & 92.20 & 89.60 & $-2.60$  \\
    \bottomrule
  \end{tabular}
\end{table}

\Cref{tab:posttraining-overall} extends the picture across the full SFT $\to$ GRPO $\to$ DPO $\to$ multi-environment GRPO pipeline on a broader benchmark mix covering reasoning (GPQA-Diamond, MMLU, MMLU-Pro), instruction-following (IFEval), tool calling (BFCL), and code (HumanEval, MBPP, LiveCodeBench).
Three patterns are visible.
First, GRPO preserves or modestly improves most non-math metrics, but produces a small regression on all four IFEval splits ($-1.4$ to $-3.9$) and on both BFCL splits ($-1.4$ non-live, $-1.8$ live); DPO partially recovers that ground without fully restoring the SFT baseline, ending within roughly two points of SFT on every IFEval split and on both BFCL splits.
Second, GPQA-Diamond and MMLU-Pro compound across stages, with GPQA-Diamond gaining $+14.1$ across the full pipeline and contributing its largest single jump at the DPO stage ($+10.1$), consistent with DPO acting on a much broader reasoning-style preference distribution than the math-only GRPO prompt set.
Third, the multi-environment GRPO stage is where IFEval finally crosses back above its SFT baseline — prompt-level strict accuracy lifts by $+5.2$ over the DPO checkpoint and ends $+3.4$ above SFT — and BFCL similarly recovers ($+0.7$ non-live, $+1.2$ live), ending within a point of the SFT baseline on both splits.

\begin{table}[!htbp]
  \caption{Benchmark scores across the full post-training pipeline (SFT $\to$ GRPO math-only $\to$ DPO $\to$ multi-env GRPO). All values are percentage accuracy; BFCL scores have been rescaled from $[0, 1]$ to $[0, 100]$ for consistency. The final column corresponds to the released v1.0 weights.}
  \label{tab:posttraining-overall}
  \centering
  \small
  \begin{tabular}{lrrrr}
    \toprule
    \textbf{Benchmark} & \textbf{SFT} & \textbf{+ GRPO} & \textbf{+ DPO} & \textbf{+ multi-env GRPO} \\
    \midrule
    GPQA-Diamond                        & 35.86 & 39.39 & 49.49 & 50.00 \\
    MMLU (weighted)                     & 76.74 & 78.04 & 79.11 & 80.30 \\
    MMLU-Pro (weighted)                 & 58.33 & 64.90 & 66.53 & 67.70 \\
    IFEval prompt-level (strict)        & 76.52 & 74.31 & 74.68 & 79.90 \\
    IFEval instruction-level (strict)   & 83.45 & 82.01 & 82.61 & 86.20 \\
    IFEval prompt-level (loose)         & 81.89 & 78.00 & 80.04 & 84.30 \\
    IFEval instruction-level (loose)    & 87.53 & 84.65 & 86.57 & 89.20 \\
    BFCL non-live                       & 76.20 & 74.80 & 75.20 & 75.90 \\
    BFCL live                           & 68.50 & 66.70 & 67.10 & 68.30 \\
    HumanEval                           & 92.68 & 93.29 & 93.90 & 96.30 \\
    MBPP                                & 74.60 & 75.60 & 74.40 & 76.80 \\
    LiveCodeBench (code generation)     & 52.75 & 54.25 & 55.50 & 55.00 \\
    \bottomrule
  \end{tabular}
\end{table}

Read together, the two tables show the role of each stage: GRPO drives the math-cluster gains; DPO compounds reasoning improvements and partially recovers the small instruction-following and tool-use regressions that GRPO introduces; multi-environment GRPO then advances most metrics — most visibly on IFEval (prompt-level strict $+5.2$, instruction-level strict $+3.6$, prompt-level loose $+4.3$, instruction-level loose $+2.6$), code generation (HumanEval $+2.4$, MBPP $+2.4$), reasoning (GPQA-Diamond $+0.5$, MMLU $+1.2$, MMLU-Pro $+1.2$), and tool calling (BFCL non-live $+0.7$, BFCL live $+1.2$), at the cost of a fractional dip on LiveCodeBench ($-0.5$).

\section{Evaluation}
\label{sec:evaluation}

\subsection{Setup}
\label{sec:evaluation:setup}

We evaluate \modelname{} against four peer models drawn from the 7--10\,B parameter class:
Qwen3.5-9B~\citep{qwen2026qwen35}, OLMo-3-7B-Think~\citep{allenai2025olmo3},
Llama-3.1-Nemotron-Nano-8B-v1~\citep{nvidia2025nemotronnano}, and
Ministral-3-8B-Reasoning~\citep{mistral2025ministral3b}.
Comparisons to models larger than 10\,B parameters are deliberately excluded: a 10\,B
reasoning model deployed in a regulated enterprise is expected to compete against models
of equivalent serving cost rather than against larger frontier systems.

All evaluations were run using the public release of each peer model, \emph{thinking on} mode and the sampling parameters recommended
by each model provider.
\modelname{}'s sequence length was 32{,}768 tokens for every benchmark except RULER, for which it was
extended to 131{,}072 (128k) tokens through YARN to accommodate the 32k and 64k retrieval probes.
This setup applies uniformly for all the suites: reasoning (MATH-500~\citep{hendrycks2021math500}, AIME~2025~\citep{maa2025aime}, GPQA-Diamond~\citep{rein2023gpqa}),
general knowledge (MMLU~\citep{hendrycks2020mmlu}, MMLU-PRO~\citep{wang2024mmlupro}), instruction following (IFEval~\citep{zhou2023ifeval}), code (HumanEval~\citep{chen2021humaneval},
LiveCodeBench~\citep{jain2024livecodebench}, MBPP~\citep{austin2021mbpp}), multilingual (MGSM~\citep{shi2022mgsm}), long-context retrieval
(RULER~32k and 64k~\citep{hsieh2024ruler}), and tool calling (BFCL~V3~\citep{patil2024bfcl}).
Math benchmarks use avg@48 sampling, where stochastic decoding is the standard reporting
convention.
Unless otherwise noted, all reported scores are percentages (\%).
\modelname{}'s dual-mode comparison closes the section.

\textbf{\modelname{} Evaluation Status.}
The \modelname{} results presented here correspond to the latest checkpoint of the
multi-environment GRPO stage (\Cref{sec:posttraining:multienv}); values may be
refreshed upon final model release.

\textbf{Token efficiency as a first-class metric.}
For deployment in regulated enterprises the relevant cost function is not pass@1 alone
but \emph{cost per correct answer}: accuracy weighted by the number of tokens generated
to reach it.
Two models with identical accuracy but a 2$\times$ token-budget difference impose
materially different inference cost, latency, and context-window pressure on downstream
agentic systems.
We therefore report mean generated tokens alongside accuracy and treat the two as joint
outputs of a deployable model.

\subsection{Efficiency Frontier}
\label{sec:evaluation:efficiency}

\begin{figure}[!htbp]
  \centering
  \includegraphics[width=0.92\textwidth]{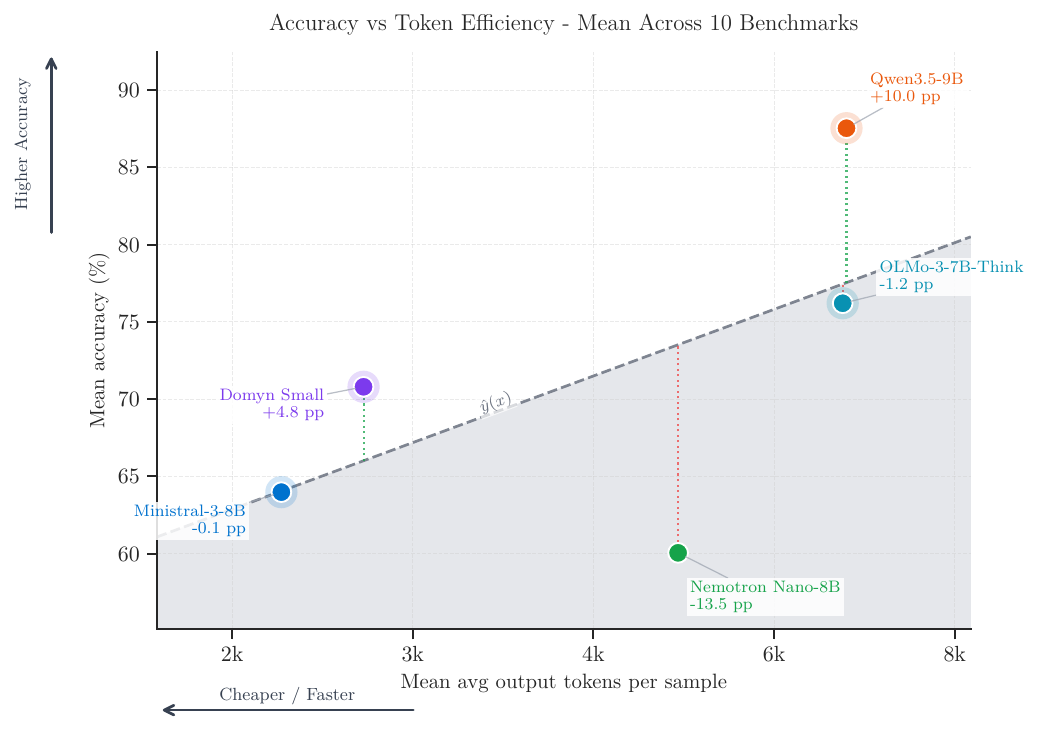}
  \caption{Accuracy/token-budget efficiency frontier across the ten benchmarks listed in \Cref{tab:tokens}.
    Each point is one model; the dashed line $\hat{y}(x)$ is the ordinary least squares (OLS) fit across the peer set.
    Models above the line deliver higher accuracy than their token budget predicts;
    models to the left are cheaper to run.}
  \label{fig:pareto_all}
\end{figure}

\modelname{}'s most distinctive result is its position on the accuracy/token-budget
Pareto frontier relative to the larger peer models, on both reasoning and code
(\Cref{fig:pareto_all}).
\Cref{tab:tokens} reports the mean number of tokens generated per problem for each model
across ten benchmarks spanning reasoning (MATH-500, AIME~2025, GPQA-Diamond), code
(HumanEval, LiveCodeBench, MBPP), general knowledge (MMLU, MMLU-PRO), instruction
following (IFEval), and multilingual (MGSM), together with per-category weighted grand means; lower values
indicate lower per-query inference cost.
While Qwen3.5-9B leads on most reasoning and instruction-following benchmarks
(\Cref{tab:capabilities}), \modelname{} achieves a superior accuracy-efficiency tradeoff
across the peer set: its reasoning token budget is approximately two thirds lower than that of
Qwen3.5-9B and OLMo-3-7B-Think, and it records the lowest token count on the
open-ended code-generation benchmark LiveCodeBench.

\begin{table}[!htbp]
  \caption{Mean generated tokens per problem across reasoning, code, general knowledge, instruction following, and multilingual benchmarks (thinking on). Lower is better. Grand means weight each benchmark by its problem count.}
  \label{tab:tokens}
  \centering
  \resizebox{\textwidth}{!}{
  \begin{tabular}{llrrrrr}
    \toprule
    Benchmark & & \modelname{} & Qwen3.5-9B & OLMo-3-7B & Nem.~Nano & Ministral-3-8B \\
    \midrule
    \multicolumn{7}{l}{\textit{Reasoning}} \\
    MATH-500   & (500 samples) & 2{,}261          & 7{,}614  & 6{,}019  & 3{,}320  & \textbf{1{,}642} \\
    AIME 2025  & (30 samples)  & 5{,}190          & 18{,}668 & 13{,}600 & 8{,}403  & \textbf{5{,}088} \\
    GPQA-D     & (198 samples) & 3{,}396          & 8{,}976  & 10{,}832 & 2{,}234  & \textbf{1{,}305} \\
    \multicolumn{2}{l}{Reasoning grand mean} & 2{,}690 & 8{,}440 & 7{,}641 & 3{,}234 & \textbf{1{,}692} \\
    \midrule
    \multicolumn{7}{l}{\textit{Code}} \\
    HumanEval  & (164 samples) & 1{,}884          & 1{,}144  & 3{,}230  & 3{,}604  & \textbf{271} \\
    LiveCodeBench    & (400 samples) & \textbf{5{,}010} & 12{,}739 & 11{,}299 & 9{,}538  & 7{,}123 \\
    MBPP       & (500 samples) & 2{,}420          & 1{,}927  & 4{,}825  & 9{,}478  & \textbf{1{,}065} \\
    \multicolumn{2}{l}{Code grand mean} & 3{,}312 & 5{,}870 & 7{,}012 & 8{,}595 & \textbf{3{,}220} \\
    \midrule
    \multicolumn{7}{l}{\textit{General Knowledge}} \\
    MMLU       & (14{,}042 samples) & 1{,}236          & 3{,}262  & 3{,}090  & 1{,}694  & \textbf{656} \\
    MMLU-PRO   & (12{,}032 samples) & 2{,}947          & 4{,}666  & 6{,}472  & 2{,}170  & \textbf{961} \\
    \multicolumn{2}{l}{General Knowledge grand mean} & 2{,}026 & 3{,}910 & 4{,}651 & 1{,}914 & \textbf{797} \\
    \midrule
    \multicolumn{7}{l}{\textit{Instruction Following}} \\
    IFEval     & (541 samples)      & 775              & 3{,}874  & 2{,}771  & \textbf{351} & 527 \\
    \midrule
    \multicolumn{7}{l}{\textit{Multilingual}} \\
    MGSM       & (250 per lang)     & 796              & 3{,}140  & 3{,}557  & 11{,}239 & \textbf{454} \\
    \bottomrule
  \end{tabular}
  }
\end{table}

\paragraph{Reasoning.}
\modelname{}'s weighted grand mean of 2{,}690 tokens per problem is approximately 32\,\%
of Qwen3.5-9B's 8{,}440 and 35\,\% of OLMo-3-7B-Think's 7{,}641, representing
3.1$\times$ and 2.8$\times$ token savings respectively.
Against Nemotron-Nano, \modelname{} is more modest at 83\,\% of its grand mean (a
roughly 17\,\% reduction); Ministral-3-8B is the only peer with a lower reasoning grand
mean (1{,}692). 
We observe that the generally low token counts of Ministral-3-8B are in part caused by a tendency of the model to answer questions within the reasoning trace (omitting the \texttt{[/THINK]} token) rather than closing it and generating a final answer.
On AIME~2025 specifically, \modelname{} produces 28\,\% of the tokens required by
Qwen3.5-9B (5{,}190 vs.\ 18{,}668) and 38\,\% of those generated by OLMo-3-7B-Think
(vs.\ 13{,}600), at comparable accuracy to Ministral-3-8B on this hard-math benchmark
(35.7 vs.\ 32.3 at 5{,}190 vs.\ 5{,}088 tokens respectively).
We attribute the reasoning-side efficiency advantage to two post-training stages: the SFT run (\Cref{sec:posttraining:sft}), which biases the supervised distribution toward shorter validated reasoning traces, and the GRPO
length-shaping reward (\Cref{sec:posttraining:grpo}), which directly penalises wasted
reasoning tokens.

\paragraph{Code.}
The efficiency picture on code is more split.
On LiveCodeBench, \modelname{} produces 5{,}010
tokens per problem, the lowest among all five peers, ahead of Ministral-3-8B (7{,}123),
Nemotron-Nano (9{,}538), OLMo-3-7B-Think (11{,}299), and Qwen3.5-9B (12{,}739).
On the shorter structured-completion benchmarks, however, Qwen3.5-9B is more
token-efficient than \modelname{}: 1{,}144 vs.\ 1{,}884 on HumanEval and
1{,}927 vs.\ 2{,}420 on MBPP.
Aggregated across the three code benchmarks, \modelname{}'s grand mean of 3{,}312
tokens lands at 56\,\% of Qwen3.5-9B's 5{,}870, 47\,\% of OLMo-3-7B-Think's 7{,}012,
and 39\,\% of Nemotron-Nano's 8{,}595; Ministral-3-8B's grand mean of 3{,}220 is
essentially tied with \modelname{} at the aggregate level, even though Ministral-3-8B is
markedly more token-efficient than \modelname{} on HumanEval and MBPP individually.

At any given accuracy threshold on reasoning tasks, \modelname{} consumes roughly
one-third the inference compute of Qwen3.5-9B (the strongest peer on raw accuracy) and
under 35\,\% that of OLMo-3-7B-Think, reducing end-to-end latency on token-bound
generation and leaving substantially more headroom in a fixed-context agentic loop.

\subsection{Capability Snapshot}
\label{sec:evaluation:capabilities}

\Cref{tab:capabilities} reports \modelname{}'s accuracy across reasoning, code,
general knowledge, instruction following, multilingual, long-context retrieval, and tool
calling alongside the four peer models.

\begin{table}[!tbp]
  \caption{Capability snapshot across \modelname{} and peer models. All scores are presented as percentages (\%). All benchmarks are reported with thinking on and provider-recommended sampling parameters; Bold denotes best in row among models with a reported score.}
  \label{tab:capabilities}
  \centering
  \resizebox{\textwidth}{!}{
  \begin{tabular}{llccccc}
    \toprule
    Category & Benchmark & \modelname{} & Qwen3.5-9B & OLMo-3-7B & Nem.~Nano & Ministral-3-8B \\
    \midrule
    \multirow{3}{*}{Reasoning}
                              & MATH-500                                          & 93.2          & \textbf{97.4} & 96.8          & 95.4  & 89.2 \\
                              & AIME 2025 (avg@48)                                & 35.7          & \textbf{90.0} & 70.4          & 51.2  & 32.3 \\
                              & GPQA-Diamond                                      & 50.0          & \textbf{82.7} & 50.8          & 42.4  & 43.9 \\
    \midrule
    \multirow{3}{*}{Code}
                              & HumanEval (pass@1)                                & \textbf{96.3} & 93.3          & 95.7          & 91.5  & 86.6 \\
                              & LiveCodeBench (pass@1)                            & 55.0          & \textbf{86.2} & 74.8          & 67.2  & 46.0 \\
                              & MBPP (pass@1)                                     & 76.8          & 76.8          & \textbf{86.6} & 77.6  & 66.6 \\
    \midrule
    \multirow{2}{*}{General Knowledge}
                              & MMLU                                              & 80.3          & \textbf{84.6} & 75.2          & 56.0  & 75.3 \\
                              & MMLU-PRO                                          & 67.7          & \textbf{84.4} & 64.0          & 28.8  & 62.0 \\
    \midrule
    Instruction Following     & IFEval (strict)                                   & 79.9          & \textbf{91.0} & 83.7          & 70.4  & 62.5 \\
    \midrule
    Multilingual              & MGSM                                              & 73.1          & \textbf{88.9} & 64.0          & 19.9  & 75.5 \\
    \midrule
    \multirow{2}{*}{Long context}
                              & RULER~32k                                         & 59.5          & \textbf{89.8} & 69.8          & 34.0  & 88.7 \\
                              & RULER~64k                                         & 29.6          & \textbf{87.9} & 17.2          & 18.7  & 85.9 \\
    \midrule
    \multirow{3}{*}{Tool calling}
                              & BFCL V3 Non-Live                                  & 75.9          & \textbf{78.1} & 61.1          & 63.3  & -- \\
                              & BFCL V3 Live                                      & 68.3          & \textbf{78.4} & 66.9          & 40.2  & -- \\
                              & BFCL V3 Multi-Turn                                &  7.0          & \textbf{50.6} &  2.1          &  0.1  & -- \\
    \bottomrule
  \end{tabular}
  }
\end{table}

\textbf{Capability summary.}
\modelname{} leads HumanEval (96.3, ahead of OLMo-3-7B-Think 95.7, Qwen3.5-9B 93.3,
Nemotron-Nano 91.5, and Ministral-3-8B 86.6) and the knowledge cluster (MMLU 80.3,
MMLU-PRO 67.7), sitting second only to Qwen3.5-9B and ahead of every other peer by
3.7 to 38.9 points.
On IFEval (79.9) it ranks third, trailing OLMo-3-7B-Think (83.7) but leading
Nemotron-Nano (70.4) and Ministral-3-8B (62.5); we read the instruction-following results as a direct
consequence of the SFT instruction mixture (\Cref{sec:posttraining:sft}) and the
instruction-following domain in the final multi-environment GRPO stage
(\Cref{sec:posttraining:multienv}).
It is at rough parity with Qwen3.5-9B on MBPP (76.8 vs.\ 76.8), with OLMo-3-7B-Think on
GPQA-Diamond (50.0 vs.\ 50.8, reached at 31\,\% of OLMo's token budget), and with
Ministral-3-8B on AIME~2025 (35.7 vs.\ 32.3).
The residual gaps concentrate on hard-math and competitive-coding reasoning --- AIME~2025
trails Qwen3.5-9B by 54.3 points, OLMo-3-7B-Think by 34.7, and Nemotron-Nano by 15.5;
LiveCodeBench trails the same three peers by 31.2, 19.8, and 12.2 --- and on the
knowledge and instruction following cluster vs.\ Qwen3.5-9B specifically ($-4.3$ MMLU, $-16.7$ MMLU-PRO,
$-11.1$ IFEval), while \modelname{} remains ahead of every other peer on MMLU and
MMLU-PRO, and ahead of Nemotron-Nano and Ministral-3-8B on IFEval.
We attribute the residual reasoning gaps to a 2024-vintage base checkpoint.

\textbf{Long context.}
On RULER, \modelname{} reaches 59.5 at 32k --- ahead of Nemotron-Nano (34.0) but behind
OLMo-3-7B-Think (69.8) and the long-context-strong Qwen3.5-9B (89.8) and Ministral-3-8B
(88.7) --- and drops to 29.6 at 64k, alongside OLMo-3-7B-Think (17.2) and Nemotron-Nano
(18.7), while Qwen3.5-9B (87.9) and Ministral-3-8B (85.9) retain most of their 32k
accuracy.
The CPT stage (\Cref{sec:pretraining:cpt}) extends the native context to 32k, and the
YaRN extrapolation used to reach 64k does not generalise as well as native long-context
training --- the strategy Qwen3.5-9B and Ministral-3-8B clearly adopted.
The 32k number itself is mid-pack in the peer set, indicating headroom on the
long-context CPT recipe even before the YaRN extension question; both observations
point to a future long-context CPT phase as the candidate fix.

\textbf{Tool calling and BFCL.}\label{sec:evaluation:bfcl}
Tool calling is one of \modelname{}'s strongest deployment-relevant capabilities.
On the BFCL~V3 single-turn splits the model reaches 75.9 Non-Live and 68.3 Live, fairly close
to Qwen3.5-9B (78.1 / 78.4) and above OLMo-3-7B-Think (61.1 / 66.9) and
Nemotron-Nano (63.3 / 40.2); it does so at 280 mean tokens per problem against 590
for Qwen3.5-9B and 2{,}429 for OLMo-3-7B-Think, the best accuracy-per-token tool-calling
profile in the peer set among models that fully engage the reasoning path.
The residual weakness is the Multi-Turn split, where \modelname{} reaches 7.0 points ---
still ahead of OLMo-3-7B-Think (2.1) and Nemotron-Nano (0.1), but far behind Qwen3.5-9B
(50.6); we attribute this to the long-context limitations described above, since the Multi-Turn split requires retaining tool-use context across multiple turns and therefore benefits from a longer effective context window.
Ministral-3-8B is excluded from the BFCL comparison: during evaluation the model
consistently failed to close the \texttt{[/THINK]} reasoning delimiter, rendering its
structured outputs unparseable by the function-calling benchmark --- a format-compliance
gap worth flagging independently, since tool-calling reliability in production depends
on deterministic output structure. Further investigation is needed and is left to future work to determine the root cause of the issue.

\textbf{Effect of the reasoning toggle.}
The thinking-on / thinking-off toggle yields the largest within-benchmark gains on
code generation and multi-step science reasoning: $+26.8$ points on HumanEval,
$+22.2$ on MBPP, $+21.2$ on LiveCodeBench, $+13.4$ on MGSM, $+10.0$ on GPQA-Diamond,
and $+7.7$ on MMLU-PRO (\Cref{tab:dualmode}).\footnote{Thinking-off numbers are from a dedicated thinking-off evaluation pass on the released checkpoint, paired with the corresponding thinking-on entries from \Cref{tab:capabilities}}
Math benchmarks that already saturate at high accuracy in thinking-off mode show
smaller gains (MATH-500 $+1.8$, AIME~2025 $+4.7$), and instruction-following moves
least (IFEval $+1.3$) --- consistent with the expectation that thinking modes help
most when the bottleneck is multi-step search or program synthesis rather than
recall or format compliance.

\begin{table}[!htbp]
  \caption{Effect of the reasoning toggle on \modelname{}. All scores are presented as percentages (\%). Thinking-off values are measured on the released checkpoint with the same evaluation harness; thinking-on AIME~2025 is reported as avg@48, all other thinking-on entries are single-pass.}
  \label{tab:dualmode}
  \centering
  \begin{tabular}{lccc}
    \toprule
    Benchmark & Thinking off & Thinking on & $\Delta$ \\
    \midrule
    MATH-500           & 91.4 & 93.2 & $+1.8$ \\
    AIME 2025          & 31.0 & 35.7 & $+4.7$ \\
    LiveCodeBench      & 33.8 & 55.0 & $+21.2$ \\
    MBPP               & 54.6 & 76.8 & $+22.2$ \\
    HumanEval          & 69.5 & 96.3 & $+26.8$ \\
    GPQA-Diamond       & 40.0 & 50.0 & $+10.0$ \\
    MMLU-PRO           & 60.0 & 67.7 & $+7.7$ \\
    MGSM               & 59.7 & 73.1 & $+13.4$ \\
    IFEval (prompt strict) & 78.6 & 79.9 & $+1.3$ \\
    \bottomrule
  \end{tabular}
\end{table}

\subsection{Domyn Swarm: Inference Infrastructure}
\label{sec:domynswarm}

All evaluations and synthetic data generation runs described in this report were orchestrated with Domyn Swarm,\footnote{\url{https://github.com/igeniusai/domyn-swarm}; Apache~2.0, initial public release 6 October 2025.} an open-source, platform-agnostic toolkit for deploying language-model serving endpoints and executing high-throughput batch inference on HPC clusters.
Domyn Swarm launches scalable vLLM endpoints behind an OpenAI-compatible API, supports multiple independent model replicas with health checking and load balancing, and provides a DataFrame-to-DataFrame batch execution layer with automatic Parquet checkpointing for fault tolerance and resumable runs.
On Leonardo, Domyn Swarm was used in two roles: (1)~serving teacher models for synthetic data generation during the SFT and RL data pipelines (\Cref{sec:posttraining:sft} and \Cref{sec:posttraining:grpo}), and (2)~running the full evaluation suite across all peer models under matched decoding parameters.
The Slurm backend uses Singularity containers and job arrays with Nginx-based load balancing, enabling consistent, reproducible inference across hundreds of GPUs without manual endpoint management.

\section{Safety and AI Act Compliance}
\label{sec:safety}

This section presents \modelname{}'s safety evaluation across hazardous knowledge recall, defensive cybersecurity capability, bias, refusal calibration, and jailbreak robustness, and frames the results in the context of the EU AI Act~\citep{eu2024aiact}.
Where possible we contextualise results against published numbers from peer models in the 7--10\,B class.

\subsection{Evaluation Setup}
\label{sec:safety:setup}

All safety benchmarks were run on the released \modelname{}-v1.0 checkpoint using the open-source Inspect AI evaluation harness~\citep{inspectai2024}, with thinking mode enabled.
The model was served as an OpenAI-compatible endpoint and decoded at $\text{temperature}=0.6$, $\text{top}_p=0.9$, $\text{top}_k=25$, with no frequency or presence penalty.
Model-graded benchmarks (XSTest and StrongREJECT) used OLMo-3.1-32B-Instruct~\citep{allenai2025olmo31} as the external scorer.
The suite covers four safety-relevant dimensions: hazardous knowledge recall (WMDP~\citep{li2024wmdp}), defensive cybersecurity capability (CyberMetric~\citep{tihanyi2024cybermetric}, SecQA~\citep{liu2024secqa}), bias (BBQ~\citep{parrish2022bbq}, StereoSet~\citep{nadeem2021stereoset}), and refusal calibration and jailbreak robustness (XSTest~\citep{rottger2024xstest}, StrongREJECT~\citep{souly2024strongreject}).
Headline metrics are summarised in~\Cref{tab:safety-headline}; per-benchmark details follow.

\begin{table}[!htbp]
  \caption{\modelname{} safety evaluation headline numbers. ``Direction'' indicates whether higher or lower values are preferable for the metric reported in the same row.}
  \label{tab:safety-headline}
  \centering
  \small
  \begin{tabular}{llrr}
    \toprule
    \textbf{Dimension} & \textbf{Benchmark / metric} & \textbf{Value} & \textbf{Direction} \\
    \midrule
    \multirow{3}{*}{Hazardous knowledge}
      & WMDP-bio (accuracy)               & 73.84\,\%  & lower better \\
      & WMDP-chem (accuracy)              & 52.94\,\%  & lower better \\
      & WMDP-cyber (accuracy)             & 50.98\,\%  & higher better \\
    \midrule
    \multirow{2}{*}{Defensive cybersecurity}
      & CyberMetric-80 (accuracy)         & 95.00\,\%  & higher better \\
      & SecQA-v1 (accuracy)               & 100.00\,\% & higher better \\
    \midrule
    \multirow{4}{*}{Bias}
      & BBQ overall (accuracy)            & 95.39\,\%  & higher better \\
      & BBQ ambiguous (accuracy)          & 94.41\,\%  & higher better \\
      & StereoSet LMS (accuracy)          & 97.40\,\%  & higher better \\
      & StereoSet SS (stereotype score)   & 56.70      & 50 = ideal \\
    \midrule
    \multirow{2}{*}{Refusal calibration}
      & XSTest safe (refusal rate)        & 8.40\,\%   & lower better \\
      & XSTest unsafe (refusal rate)      & 79.50\,\%  & higher better \\
    \midrule
    \multirow{2}{*}{Jailbreak robustness}
      & StrongREJECT (jailbreak rate)     & 12.78\,\%  & lower better \\
      & StrongREJECT (composite metric)   & 0.6246     & lower better \\
    \bottomrule
  \end{tabular}
\end{table}

\subsection{Hazardous Knowledge}
\label{sec:safety:wmdp}

WMDP~\citep{li2024wmdp} measures multiple-choice accuracy on hazard-proxy questions in biosecurity, chemistry, and cybersecurity.
Following the convention of the original benchmark, lower accuracy on the biosecurity and chemistry subsets is desirable (less recall of weaponisable knowledge), while higher accuracy on the cybersecurity subset is desirable (defensive knowledge useful to practitioners).
\modelname{} scores 73.84\,\% on WMDP-bio, 52.94\,\% on WMDP-chem, and 50.98\,\% on WMDP-cyber.
These numbers are consistent with published results from general-purpose models of comparable scale trained on large web mixtures: for reference, the original WMDP paper reports WMDP-bio accuracy in the 64--78\,\% range for 7--13\,B models without targeted unlearning~\citep{li2024wmdp}.
No post-hoc unlearning procedure was applied to the released checkpoint; these results reflect the pre-existing knowledge distribution of the base model.

\subsection{Defensive Cybersecurity}
\label{sec:safety:cyber}

We complement WMDP-cyber with two defensive-cybersecurity question-answering benchmarks.
On CyberMetric-80~\citep{tihanyi2024cybermetric}, an 80-question multiple-choice probe of general cybersecurity knowledge, \modelname{} reaches 95.00\,\% accuracy.
On SecQA-v1~\citep{liu2024secqa}, a 110-question security-knowledge benchmark released under CC-BY-NC-SA~4.0 and used here for evaluation purposes only, the model reaches 100\,\%.
Both results indicate strong baseline defensive-cyber knowledge, in line with what comparable 7--10\,B models achieve on these benchmarks at saturation.

\subsection{Bias}
\label{sec:safety:bias}

\textbf{BBQ.}
BBQ~\citep{parrish2022bbq} pairs each question with either an \emph{ambiguous} context (the correct answer is ``unknown'' / ``cannot be determined'') or a \emph{disambiguated} context (the correct answer is identifiable from the passage).
\modelname{} reaches 95.39\,\% overall accuracy across all 11 BBQ categories (58{,}492 samples), with 94.41\,\% on the ambiguous condition and 96.38\,\% on the disambiguated condition.
The ambiguous-condition score is the more diagnostic: it measures whether the model resists defaulting to a stereotype when the evidence is insufficient, and 94.41\,\% places \modelname{} in the upper range for models at this scale.

\textbf{StereoSet.}
StereoSet~\citep{nadeem2021stereoset} jointly measures language-modelling fluency (LMS, higher is better) and the tendency to prefer stereotypical over anti-stereotypical continuations (Stereotype Score SS, where 50 is the unbiased ideal).
The fluency component is essentially saturated.
The SS of 56.70 indicates a mild lean toward stereotypical continuations, consistent with the range typically observed in 7--10\,B models trained on large web corpora; for comparison, the original StereoSet paper reports SS values of 56--62 for models in this parameter band~\citep{nadeem2021stereoset}.

\subsection{Refusal Calibration}
\label{sec:safety:refusal}

XSTest~\citep{rottger2024xstest} probes the trade-off between over-refusal and under-refusal by pairing 250 \emph{safe} prompts (questions that superficially resemble dangerous queries but have benign answers) with 200 \emph{unsafe} prompts (questions that the model is expected to refuse).
On the safe subset \modelname{} refuses 8.40\,\% of prompts, indicating low over-refusal --- the model correctly answers the vast majority of benign queries that could trigger false positives.
On the unsafe subset the refusal rate is 79.50\,\%, meaning the model correctly refuses four out of five genuinely harmful requests.
This is a reasonable baseline for a general-purpose 10\,B model that has not undergone a dedicated safety-tuning stage, and leaves headroom for improvement through targeted refusal training in subsequent iterations.

\subsection{Jailbreak Robustness}
\label{sec:safety:jailbreak}

StrongREJECT~\citep{souly2024strongreject} probes a model's behaviour under 313 jailbreak-style prompts with harmful intent and reports two metrics from a model-graded scorer: a \emph{jailbreak success rate} (the fraction of prompts that elicited a harmful, on-topic completion; lower is better) and a composite \emph{StrongREJECT score} in $[0, 1]$ that combines refusal, refusal quality, and harmfulness (lower is better).
\modelname{} reaches a jailbreak success rate of 12.78\,\% and a StrongREJECT composite of 0.6246.
The jailbreak success rate is competitive with general-purpose models at this scale that have not undergone dedicated adversarial training, and the composite score reflects an overall profile where the majority of harmful prompts are either refused or produce low-quality completions.

\subsection{Summary}

Across the safety suite, \modelname{} demonstrates solid performance on bias (BBQ 95.4\,\%, StereoSet ICAT 84.4), defensive cybersecurity (CyberMetric 95.0\,\%, SecQA 100\,\%), and low over-refusal (XSTest safe 8.4\,\%), with results consistently in line with or above peers at the 7--10\,B scale.
Unsafe-refusal calibration and jailbreak robustness are identified as the most natural candidates for improvement through a dedicated safety-tuning stage in future iterations.

\subsection{EU AI Act Compliance}
\label{sec:safety:aiact}

\modelname{} is released as a general-purpose AI (GPAI) model under the EU AI Act~\citep{eu2024aiact}.
Article 53 of the Act places three substantive transparency obligations on providers of GPAI models: (i) maintain up-to-date technical documentation of the model; (ii) provide information and documentation to downstream providers integrating the model into AI systems; and (iii) publish a sufficiently detailed summary of the content used for training.
This report, together with the public model card and the MIT-licensed open-weights release, is intended to discharge those obligations: the architecture, training data composition, training stages, evaluations, and known limitations are documented end-to-end, and the training-data summary required by Article 53(1)(d) is provided as a companion artefact to the model release.
To uphold data subject rights and comply with the AI Act and EU copyright framework, we operate an opt-out procedure for rights holders.
Anyone who believes their copyrighted material was inadvertently included in our training corpora can contact \texttt{copyright@domyn.com}, and we will exclude the affected data from subsequent model iterations.

\section{Conclusion}
\label{sec:contributions}

We presented \modelname{}, a 10-billion-parameter open-weight reasoning language model trained end-to-end on European HPC infrastructure and released under the MIT license.
Starting from a 2024-vintage foundation checkpoint, we showed that a calibrated sequence of continued pre-training, supervised fine-tuning, and multi-stage reinforcement learning can yield a model that achieves the strongest accuracy-efficiency balance in the 7--10\,B class: \modelname{} produces roughly one-third the tokens of Qwen3.5-9B on core reasoning benchmarks while maintaining competitive instruction-following, science reasoning, and tool-calling capability.

We documented the full adaptation pipeline at recipe-level fidelity --- CPT data composition and context-extension strategy, SFT mixture design, math-only GRPO with verifiable rewards, DPO under the Delta Learning Hypothesis, and multi-environment GRPO across five task domains --- so that each stage can be inspected and reused across new base models and domains.
Alongside the model, we release Domyn Swarm (\Cref{sec:domynswarm}), an open-source inference toolkit that supported all evaluation and synthetic data generation runs in this work.

Residual gaps on hard-math reasoning and long-context retrieval beyond 32k are documented throughout the report and inform the roadmap below.

\modelname{} is part of a broader effort to build sovereign-EU AI for regulated industries.
Future iterations will continue to treat inference efficiency as a first-class design objective, deepen multilingual European coverage to strengthen linguistic and cultural alignment, and extend enterprise-domain capabilities in financial services, defence, and advanced manufacturing through targeted continued pre-training.
A central priority is agentic AI: improving multi-turn tool calling, long-horizon planning, and autonomous task execution so that Domyn models can serve as reliable, auditable agents in production workflows where data residency and governance are non-negotiable.
All development will remain transparent and aligned with the EU AI Act, consistent with our commitment to delivering capable, governable AI under European regulatory frameworks.

We release this report to serve both the research community evaluating post-training methodology and the regulated enterprises considering deployment of open-weight reasoning models in sovereign EU settings.

\section*{Acknowledgments}
\label{sec:acknowledgments}

We gratefully acknowledge CINECA for hosting and operating the Leonardo supercomputer~\citep{turisini2024leonardo} and for the computational allocation that made the full adaptation pipeline described in this work possible.
We thank the CINECA technical and support staff for their continued assistance throughout the training and evaluation campaigns.
We thank NVIDIA for the DGX Cloud partnership that enabled the foundation pre-training of Italia~10B and for continued NeMo, NeMo~RL, and Megatron-LM framework support throughout the training phases.

\newpage
\bibliography{references}
\bibliographystyle{iclr2026_conference}

\end{document}